\documentclass[twoside,11pt]{article}

\usepackage{automl2019}
\usepackage{amsmath}
\usepackage{lscape} 
\usepackage{adjustbox} 
\usepackage[section]{placeins}

\usepackage{algpseudocode} 
\usepackage{algorithm}
\usepackage{multirow}

\usepackage[
	textsize=footnotesize,
]{todonotes}

\DeclareMathOperator*{\argmin}{\arg\min}

\jmlrheading{W. Shao and C. Gei\ss ler}

\ShortHeadings{Graduated Optimisation of Black-Box Function}{Shao and Gei\ss ler}
\firstpageno{1}

\begin{document}

\title{Graduated Optimization of Black-Box Functions}

\author{\name Weijia Shao \email weijia.shao@campus.tu-berlin.de \\
       \AND
       \name Christian Gei\ss ler \email christian.geissler@gt-arc.com \\
       \AND
       \name Fikret Sivrikaya \email fikret.sivrikaya@gt-arc.com \\ \\
       }

\maketitle

\begin{abstract}
Motivated by the problem of tuning hyperparameters in machine learning, we present a new approach for gradually and adaptively optimizing an unknown function using estimated gradients. We validate the empirical performance of the proposed idea on both low and high dimensional problems. The experimental results demonstrate the advantages of our approach for tuning high dimensional hyperparameters in machine learning.  
\end{abstract}

\section{Introduction}
Machine learning applications and the design of complex systems usually involve a large number of free parameters. The evaluation of a single set of parameters requires computationally expensive numerical simulations and cross-validations, while the choices of parameters influence the performance of a system dramatically. In the machine learning community, this problem is usually referred to as hyperparameter optimization (\textbf{HPO})\citep{hutter2011sequential} and has been extensively studied in recent years, since the early approaches based on grid search become impractical for high dimensional hyperparameters \citep{franceschi2017forward}. 

In this paper, we focus on continuous parameters, for which gradient based methods \citep{maclaurin2015gradient,luketina2016scalable,pedregosa2016hyperparameter,franceschi2017forward,wu2017bayesian} have attracted attention for its fast convergence. In machine learning applications, the objective of \textbf{HPO} is usually to optimize a validation function evaluated at a stationary point of the training objective, and the gradient of the validation function can be derived from the iterative training procedure \citep{maclaurin2015gradient,luketina2016scalable}. The exact computation of the gradient is the major bottleneck, since it is computationally inefficient and has high space requirements. \cite{pedregosa2016hyperparameter} proposed the idea of approximating the gradient based on the stationary condition of the training procedure, and managed to efficiently optimize a large number of hyperparameters.

However, there are still several issues to be addressed. First of all, the gradient approximation proposed by \cite{pedregosa2016hyperparameter} relies on regularity conditions such as the stationary condition for a minimizer of the training loss, which obviously does not hold if we apply early stopping. Furthermore, the approximation is based on the assumption of the smoothness of the objective function, which is too strong in practice. Finally, the algorithms proposed by \cite{pedregosa2016hyperparameter}, \cite{maclaurin2015gradient} and \cite{franceschi2017forward} require hyperparameters such as learning rate, which become \textit{hyper hyperparameters} in \textbf{HPO}. In their experiments, those hyperparameters are manually adjusted. In practice, devising procedures adaptive to them are needed. 

In this paper, we formalize \textbf{HPO} as a problem of optimizing the output value of an unknown function. We propose an alternate idea based on the two-point estimation of the gradient \citep{nesterov2017random}, the graduated optimization \citep{hazan2016graduated} and the scale free online learning algorithm \citep{orabona2018scale}. Compared to \citep{pedregosa2016hyperparameter}, we do not assume the smoothness or any regularity conditions of the objective function. To avoid introducing further hyper hyperparameters, we apply a simple online gradient descent with an adaptive learning rate. We compare our algorithms against the state-of-the art global optimization algorithms on machine learning problems. The rest of the paper is organized as follows. In section 2, we introduce the problem setting, describe our idea of estimating the gradient and propose the algorithm. In section 3, we present the empirical performance of our algorithm. Section 4 concludes our work with some future research directions.

\section{Problem Setting and Algorithm}
Let $f: \mathcal{X}\to \mathbb{R}$ be a function defined on a compact and convex set $\mathcal{X}\subset \mathbb{R}^d$. Finding the global minimum $x^*=\arg\min_{x\in\mathcal{X}}f(x)$ is challenging in general due to non-convexity, unknown smoothness and possible noisy evaluations of the function. In the context of machine learning, $f$ returns the score, such as the cross-validation error, for a given configuration of hyperparameters selected from $\mathcal{X}$. We follow the standard procedure from the literature on global optimization \citep{hutter2011sequential,bubeck2011x,munos2011optimistic,malherbe2017global}, which attempts to minimize the function by sequentially exploring the space $\mathcal{X}$ using a finite budget of evaluations. Formally, we wish to find a sequence $x_1,\ldots, x_T$, where each point $x_t$ depends on the previous evaluations $f(x_1),\ldots,f(x_{t-1})$, such that the last explored point $x_T$ returns a lowest possible value. 

The global optimization methods do not or cannot usually leverage the gradient information, which is actually proven to be useful for tuning hyperparameters in machine learning \citep{maclaurin2015gradient,pedregosa2016hyperparameter,franceschi2017forward}. However, deriving the gradients in those works is expensive, require strong assumptions on $f$, and is not applicable to black-box problems. In contrast, our idea works for more general cases. Assuming $f$ is $L$-Lipschitz, its \textit{Gaussian approximation} \citep{nesterov2017random} is defined as
\[
f_\delta(x)= \mathbf{E}_{u}[f(x+\delta u)],
\]
where $u\sim \mathcal{N}(0,I_d)$ is a standard Gaussian random vector. $f_\delta(x)$ is $\frac{\sqrt{d}L}{\delta}$-smooth \citep{nesterov2017random} with bounded bias $|f_{\delta}(x)-f(x)|\leq \delta L$ \citep{hazan2016graduated}. The gradient $\triangledown f_\delta(x)$ can be estimated using the two-point feedback \citep{nesterov2017random}
\[
\frac{d}{\delta}(f(x+\delta u)-f(x))u.
\]
Arguably, one can estimate the gradient with only one expensive function call \citep{hazan2016graduated} or use other two-point estimators with lower variance \citep{shamir2017optimal}. Our choice is more practical, since the evaluation at each $f(x_t)$ could help us trace the best configuration of the parameters evaluated.   

\begin{algorithm}
	\caption{GradOpt}\label{GradOpt}
	\begin{algorithmic}[1]
		\State input: budget $T$, $\mathcal{X}$
		\State initialise $x_1$, $t=1$, $\delta_1=\frac{1}{2}\max_{x,y \in \mathcal{X}}\lVert x-y \rVert_2$, $T_1,\ldots,T_M$ such that $\sum_{m=1}^{M}T_m= T$
		\For{$m=1$ to $M$}
		\For{$t_m=1$ to $T_m$}
		\State Sample $u$ from $\mathcal{N}(0,I_d)$
		\State $g_t=\frac{d}{\delta_m}(f(x_t+\delta_m u)-f(x_t))u$
		\For{$i=1$ to $d$}  \Comment{scale free online gradient descent}
		\State $\eta_{i}= \eta_{i}+g_{t,i}^2$
		\State $\tilde x_{t+1,i}=x_{t,i}-\frac{g_{t,i}}{\sqrt{\eta_{i}}}$\Comment{simply do nothing if learning rate is $0$}
		\EndFor
		\State $x_{t+1}=\Pi_\mathcal{X}(\tilde x_{t+1})$ \Comment{project it to the decision space}
		\State $t=t+1$
		\EndFor
		\State $\delta_{m+1}=\frac{\delta_m}{2}$
		\EndFor
		\State return $\max_{t=1,\ldots,T}x_t$
	\end{algorithmic}
\end{algorithm}
\textbf{Algorithm \ref{GradOpt}} describes our idea. We can divide it into $M$ epochs. In each epoch, we use online gradient descent \citep{orabona2018scale} with adaptive learning rate to optimise $f_{\delta_m}$. If the smoothed functions $f_{\delta_m}$ are locally strongly convex and the global optima of smoothed functions of the successive epochs are close, a point close to the global optimum can be ensured \citep{hazan2016graduated} for a large enough budget. Otherwise it converges to a stationary point of $f$ as we gradually decrease $\delta$ \citep{nesterov2017random}.  Compared to the standard gradient descent method used in \citep{pedregosa2016hyperparameter}, our algorithm does not assume the $L$-smoothness of $f$ and derive the learning rate from $L$, which is unknown for most cases. Unlike those methods employed in \citep{maclaurin2015gradient,franceschi2017forward}, we do not use momentum term in the gradient descent to avoid additional hyperparameters.

\section{Evaluation}
In this section, we compare the empirical performance of \textbf{GradOpt} with the following global optimisation algorithms:
\begin{itemize}
	\item \textbf{PRS}. The Pure Random Search methods samples parameters uniformly randomly from the searching space.  
	\item \textbf{HOO}. The tree based global optimisation methods for H\"older continuous functions \citep{bubeck2011x}.
	\item \textbf{AdaLipo}. The sequential strategy for optimising unknown Lipschitz continuous functions while adaptively estimating the Lipschitz constant \citep{malherbe2017global}.
	\item \textbf{AdaLipoTR}. The practical method combining AdaLipo \citep{malherbe2017global} for global search, and trust region for finding the local results \footnote{The implementation of both AdaLipo and AdaLipoTR is taken from the \textbf{dlib} library (http://blog.dlib.net/2017/12/a-global-optimization-algorithm-worth.html)}.

\end{itemize}
We adopt the experimental setting of \citep{malherbe2016ranking,malherbe2017global} and consider the problems of tuning both low dimensional and high dimensional hyperparameters in machine learning. For the low dimensional case, we tune the $L_2$-regularizer and the width of a Gaussian kernel ridge regression. The objective is to maximize the 10-fold cross validation score. More specifically, we split the dataset $\{(X_i, Y_i)\}_{i=1}^n$ into 10 folds $D_1, ...D_{10}$ and consider the following objective function
\begin{equation}
1 - \frac{1}{10}\sum_{k=1}^{10}  \frac{\sum_{i\in D_k}(\hat{f}_k(X_i)-Y_i)^2}{\sum_{i\in D_k}(\overline{Y}-Y_i)^2}
\end{equation}
subject to \[\hat{f}_k \in \argmin_{f \in \mathcal{H}_{\sigma}} \frac{1}{n- |D_k|} \sum_{i \not\in D_k}(f(X_i)-Y_i)^2 + 10^\lambda \|f\|_{\mathcal{H}_{\sigma}},\]
where $\mathcal{H}_{\sigma}$ denotes the Gaussian \textbf{RKHS} of bandwidth $10^\sigma$ equipped with the norm $\|\cdot\|_{\mathcal{H}_{\sigma}}$. The goal is to search for the optimal $\lambda$ and $\sigma$ from $[-2,4]\times[-5,5]$.

To compare the performance for high dimensional hyperparameters, we consider the task of data cleaning for kernel ridge regression, for which we assign a weight from $[0,1]$ to each data sample. Then we tune the hyperparameters and weights, i.e. to maximize (1) subject to \[\hat{f}_k \in \argmin_{f \in \mathcal{H}_{\sigma}} \frac{1}{n- |D_k|} \sum_{i \not\in D_k}\mathbf{w}_i(f(X_i)-Y_i)^2 + 10^\lambda \|f\|_{\mathcal{H}_{\sigma}}.\] The decision space for $\lambda$, $\sigma$ and $\mathbf{w}_i$ for $i\in\{1,\ldots,n\}$ is $[-2,4]\times[-5,5]\times[0,1]^n$. Note that, for the convenience of applying \textbf{HOO} and \textbf{AdaLipo}, all the problems are implemented as maximization problems. To apply \textbf{GradOpt} we simply minimise the negative of the objective function. 

For each of the problems, we perform $100$ runs of the algorithms with a budget of $B=1000$ function evaluations. Then for each target value in $\{0.90,0.95,0.99\}$, we observe the number of function evaluations required to reach the best score found by the algorithms multiplied by the target value.

\begin{table}
	\begin{adjustbox}{max width=\textwidth}
\begin{tabular}{l c c c c c}
90\% Target & \textbf{Housing} & \textbf{Yacht} & \textbf{Slump} & \textbf{BreastCancer} & \textbf{AutoMPG}\\
\hline
PRS & $332.13(\pm 225)$ & $32.33(\pm 30)$ & $363.41(\pm 296)$ & $20.12(\pm 16)$ & $374.04(\pm 283)$\\
HOO & $94.27(\pm 10)$ & $\textbf{7.56}(\pm 0)$ & $1000.0(\pm 0)$ & $19.5(\pm 4)$ & $118.71(\pm 5)$\\
AdaLipo & $8.1(\pm 5)$ & $12.4(\pm 25)$ & $12.46(\pm 10)$ & $98.48(\pm 160)$ & $9.48(\pm 15)$\\
AdaLipoTR & $\textbf{6.03}(\pm 2)$ & $9.35(\pm 16)$ & $\textbf{6.74}(\pm 2)$ & $23.25(\pm 51)$ & $\textbf{5.38}(\pm 1)$\\
GradOpt & $10.38(\pm 6)$ & $16.9(\pm 22)$ & $14.41(\pm 9)$ & $\textbf{18.59}(\pm 19)$ & $8.52(\pm 5)$\\
\hline
\end{tabular}
\end{adjustbox}
\\\begin{adjustbox}{max width=\textwidth}
\begin{tabular}{l c c c c c}
95\% Target & \textbf{Housing} & \textbf{Yacht} & \textbf{Slump} & \textbf{BreastCancer} & \textbf{AutoMPG}\\
\hline
PRS & $535.49(\pm 373)$ & $43.12(\pm 43)$ & $601.05(\pm 339)$ & $31.33(\pm 26)$ & $603.8(\pm 327)$\\
HOO & $109.94(\pm 11)$ & $\textbf{7.56}(\pm 0)$ & $1000.0(\pm 0)$ & $38.79(\pm 8)$ & $133.81(\pm 6)$\\
AdaLipo & $11.45(\pm 9)$ & $13.44(\pm 25)$ & $17.4(\pm 14)$ & $106.26(\pm 160)$ & $13.27(\pm 18)$\\
AdaLipoTR & $\textbf{6.21}(\pm 2)$ & $10.26(\pm 16)$ & $\textbf{6.95}(\pm 2)$ & $24.23(\pm 51)$ & $\textbf{5.51}(\pm 1)$\\
GradOpt & $10.47(\pm 6)$ & $20.21(\pm 25)$ & $14.69(\pm 9)$ & $\textbf{19.67}(\pm 20)$ & $8.98(\pm 5)$\\
\hline
\end{tabular}
\end{adjustbox}
\\\begin{adjustbox}{max width=\textwidth}
\begin{tabular}{l c c c c c}
99\% Target & \textbf{Housing} & \textbf{Yacht} & \textbf{Slump} & \textbf{BreastCancer} & \textbf{AutoMPG}\\
\hline
PRS & $771.22(\pm 357)$ & $208.23(\pm 137)$ & $936.42(\pm 172)$ & $70.69(\pm 88)$ & $918.51(\pm 179)$\\
HOO & $608.82(\pm 301)$ & $62.0(\pm 10)$ & $1000.0(\pm 0)$ & $38.79(\pm 8)$ & $412.59(\pm 60)$\\
AdaLipo & $36.31(\pm 25)$ & $22.45(\pm 30)$ & $45.38(\pm 26)$ & $132.36(\pm 168)$ & $35.08(\pm 25)$\\
AdaLipoTR & $\textbf{6.42}(\pm 2)$ & $\textbf{12.93}(\pm 16)$ & $\textbf{7.15}(\pm 2)$ & $\textbf{26.61}(\pm 51)$ & $\textbf{5.71}(\pm 1)$\\
GradOpt & $10.7(\pm 6)$ & $43.26(\pm 37)$ & $18.17(\pm 14)$ & $34.13(\pm 49)$ & $9.54(\pm 6)$\\
\hline
\end{tabular}
\end{adjustbox}
\\
	\caption{Results of the numerical experiments on hyperparameter tuning for Gaussian kernel ridge regression}
	\label{tab:exp_ml}
\end{table}

\begin{table}
	\begin{adjustbox}{max width=\textwidth}
\begin{tabular}{l c c c c c}
90\% Target & \textbf{HousingHD} & \textbf{YachtHD} & \textbf{SlumpHD} & \textbf{BreastCancerHD} & \textbf{AutoMPGHD}\\
\hline
PRS & $247.16(\pm 218)$ & $9.17(\pm 8)$ & $997.08(\pm 29)$ & $153.39(\pm 161)$ & $609.24(\pm 373)$\\
HOO & $1000.0(\pm 0)$ & $\textbf{1.66}(\pm 1)$ & $1000.0(\pm 0)$ & $1000.0(\pm 0)$ & $1000.0(\pm 0)$\\
AdaLipo & $216.29(\pm 211)$ & $13.51(\pm 12)$ & $939.87(\pm 183)$ & $192.54(\pm 245)$ & $626.59(\pm 327)$\\
AdaLipoTR & $190.62(\pm 182)$ & $10.81(\pm 9)$ & $887.59(\pm 279)$ & $\textbf{118.4}(\pm 97)$ & $340.91(\pm 161)$\\
GradOpt & $\textbf{41.16}(\pm 50)$ & $3.21(\pm 9)$ & $\textbf{824.29}(\pm 274)$ & $145.78(\pm 293)$ & $\textbf{79.34}(\pm 60)$\\
\hline
\end{tabular}
\end{adjustbox}
\\\begin{adjustbox}{max width=\textwidth}
\begin{tabular}{l c c c c c}
95\% Target & \textbf{HousingHD} & \textbf{YachtHD} & \textbf{SlumpHD} & \textbf{BreastCancerHD} & \textbf{AutoMPGHD}\\
\hline
PRS & $704.38(\pm 356)$ & $\textbf{12.25}(\pm 11)$ & $1000.0(\pm 0)$ & $810.72(\pm 326)$ & $969.85(\pm 140)$\\
HOO & $1000.0(\pm 0)$ & $1000.0(\pm 0)$ & $1000.0(\pm 0)$ & $1000.0(\pm 0)$ & $1000.0(\pm 0)$\\
AdaLipo & $673.8(\pm 364)$ & $19.69(\pm 17)$ & $989.07(\pm 81)$ & $909.7(\pm 233)$ & $947.59(\pm 192)$\\
AdaLipoTR & $494.81(\pm 208)$ & $15.29(\pm 13)$ & $994.74(\pm 52)$ & $657.57(\pm 347)$ & $561.52(\pm 266)$\\
GradOpt & $\textbf{139.6}(\pm 101)$ & $74.38(\pm 88)$ & $\textbf{943.3}(\pm 167)$ & $\textbf{342.31}(\pm 329)$ & $\textbf{302.46}(\pm 172)$\\
\hline
\end{tabular}
\end{adjustbox}
\\\begin{adjustbox}{max width=\textwidth}
\begin{tabular}{l c c c c c}
99\% Target & \textbf{HousingHD} & \textbf{YachtHD} & \textbf{SlumpHD} & \textbf{BreastCancerHD} & \textbf{AutoMPGHD}\\
\hline
PRS & $1000.0(\pm 0)$ & $\textbf{52.28}(\pm 57)$ & $1000.0(\pm 0)$ & $1000.0(\pm 0)$ & $1000.0(\pm 0)$\\
HOO & $1000.0(\pm 0)$ & $1000.0(\pm 0)$ & $1000.0(\pm 0)$ & $1000.0(\pm 0)$ & $1000.0(\pm 0)$\\
AdaLipo & $1000.0(\pm 0)$ & $66.03(\pm 71)$ & $1000.0(\pm 0)$ & $1000.0(\pm 0)$ & $994.74(\pm 52)$\\
AdaLipoTR & $1000.0(\pm 0)$ & $59.46(\pm 59)$ & $1000.0(\pm 0)$ & $1000.0(\pm 0)$ & $\textbf{902.81}(\pm 201)$\\
GradOpt & $\textbf{981.38}(\pm 74)$ & $358.12(\pm 176)$ & $\textbf{997.83}(\pm 22)$ & $\textbf{960.94}(\pm 116)$ & $980.56(\pm 90)$\\
\hline
\end{tabular}
\end{adjustbox}
\\
	\caption{Results of the numerical experiments on high dimensional hyperparameter tuning for Gaussian kernel ridge regression (additionally tuning the training weights)}
	\label{tab:exp_mlhd}
\end{table}

\textbf{Table \ref{tab:exp_ml} and \ref{tab:exp_mlhd}} demonstrate the experimental results.\footnote{The source code can be fetched from https://github.com/christiangeissler/gradoptbenchmark} Despite the potentially suboptimal stationary points, \textbf{GradOpt} returns a point close to the optimal solutions found by other global optimisation algorithms in all experiments. For the low dimensional problems, the combined approach \textbf{AdaLipoTR} outperforms the other algorithms. However, \textbf{GradOpt} takes only a few more steps compared to \textbf{AdaLipoTR} and outperforms the algorithms relying on global search for all datasets except \textbf{Yacht}. The experimental results for the high dimensional tasks demonstrate the advantage of \textbf{GradOpt}, which obtains most of the best scores for all target values. The other global optimization algorithms don't scale well for the high dimensional problems, which is also suggested by their theoretical analysis \citep{bubeck2011x,malherbe2017global}.

\section{Discussion and Future Work}
We presented an alternative approach for optimizing an unknown, potentially non-convex and non-smoothed function, which is based on the estimated gradient, adaptive learning rate and graduated optimization. Suggested by the theoretical analysis of the previous work, our approach converges to a stationary point for general cases and to a global optimum if certain conditions are fulfilled. The experimental results have shown that our approach indeed provides global guarantee. For tuning high dimensional hyperparameters, it outperforms the state-of-the-art global optimization algorithms in most of the experiments.
 
We consider this work as a glimpse of applying graduated optimization to searching for optima of unknown functions. It can be extended and improved in several ways. Firstly, the convergence of our approach is suggested by previous work and empirically shown, yet the actual theoretical performance is unknown. The most important future direction would be to perform a theoretical analysis in an appropriate framework. Furthermore, we assign an equal budget of evaluations to each epoch in this work, which may not be best option. A strategy of allocating budget with theoretical guarantee would be needed in practice. Finally, the experimental results show that the combined method outperforms the rest for the low dimensional problems. To apply our approach to tuning hyperparameters in machine learning, we can also combine it with global optimization algorithms. However, this must be thoroughly evaluated on diverse datasets with different models.

\acks{This work is supported in part by the German Federal
Ministry of Education and Research (BMBF) under the grant
number 01IS16046. 
We would like to thank Dr. Brijnesh Jain for his valuable feedback.}

\vskip 0.2in
\bibliography{bibliography}

\end{document}